\documentclass{article}

\usepackage{microtype}
\usepackage{graphicx}
\usepackage{subfigure}
\usepackage{amsmath}
\usepackage{amsthm}
\usepackage{amsfonts}
\usepackage{stfloats}
\usepackage{multirow}
\usepackage{float}
\usepackage{enumitem}
\usepackage{booktabs} % for professional tables

\usepackage{hyperref}

\usepackage{xcolor}
\newcommand{\xhdr}[1]{\noindent{\bfseries #1}.}

\newcommand{\name}{GraIL}
\newcommand{\cut}[1]{}
\newcommand{\mb}{\mathbf}
\newtheorem{theorem}{Theorem}
\newtheorem{corollary}{Corollary}
\newtheorem{lemma}{Lemma}

\usepackage[accepted]{icml2020_arxiv}

\icmltitlerunning{Inductive Relation Prediction by Subgraph Reasoning}

\begin{document}

\twocolumn[
\icmltitle{Inductive Relation Prediction by Subgraph Reasoning}



\icmlsetsymbol{equal}{*}

\begin{icmlauthorlist}
\icmlauthor{Komal K. Teru}{mcgill,mila}
\icmlauthor{Etienne Denis}{mcgill,mila}
\icmlauthor{William L. Hamilton}{mcgill,mila}
\end{icmlauthorlist}

\icmlaffiliation{mcgill}{McGill University}
\icmlaffiliation{mila}{Mila}

\icmlcorrespondingauthor{Komal K. Teru}{komal.teru@mail.mcgill.ca}

\icmlkeywords{Machine Learning, ICML}

\vskip 0.3in
]



\printAffiliationsAndNotice{}  % leave blank if no need to mention equal contribution

\begin{abstract}
The dominant paradigm for relation prediction in knowledge graphs involves learning and operating on latent representations (i.e., embeddings) of entities and relations. 
%These methods exploit the local connectivity patterns and homophily in the knowledge graph to make predictions. 
However, these embedding-based methods do not explicitly capture the compositional logical rules underlying the knowledge graph,
% However, it is not clear if embedding-based methods effectively capture the relational semantics---i.e., the logical rules that hold among the relations underlying the knowledge graph. 
and they are limited to the transductive setting, where the full set of entities must be known during training. 
% Here, we propose a graph neural network approach, \name, for relation prediction that reasons over local subgraph structures and has a strong inductive bias to learn entity-independent relational semantics. 
Here, we propose a graph neural network based relation prediction framework, \name, that reasons over local subgraph structures and has a strong inductive bias to learn entity-independent relational semantics. 
Unlike embedding-based models, \name\ is naturally inductive and can generalize to unseen entities and graphs after training.
We provide theoretical proof and strong empirical evidence that \name\ can represent a useful subset of first-order logic and show that \name\ outperforms existing rule-induction baselines in the inductive setting. 
We also demonstrate significant gains obtained by ensembling \name\ with various knowledge graph embedding methods in the transductive setting, highlighting the complementary inductive bias of our method.
\end{abstract}

\cut{
\begin{abstract}
Inferring missing edges in multi-relational knowledge graphs is a fundamental task in statistical relational learning. However, previous work has largely focused on the transductive relation prediction problem, where missing edges must be predicted for a single, fixed graph. In contrast, many real-world situations require relation prediction on dynamic or previously unseen knowledge graphs (e.g., for question answering, dialogue, or e-commerce applications). Here, we develop a novel graph neural network (GNN) architecture to perform inductive relation prediction and provide a systematic comparison between this GNN approach and rule induction baselines. Our results highlight the significant difficulty of inductive relational learning—compared to the transductive case—and offer a new challenging set of inductive benchmarks for knowledge graph completion. We also demonstrate significant gains obtained by ensembling our method with various knowledge graph embedding methods in the transductive setting, hence proving the complementary inductive bias of our method.
\end{abstract}
}

\section{Introduction}
\label{sec:intro}

% \todo{Introduce the name \name\ more naturally.}

Knowledge graphs (KGs) are a collection of facts which specify relations (as edges) among a set of entities (as nodes). Predicting missing facts in KGs---usually framed as relation prediction between two entities---is a widely studied problem in statistical relational learning \cite{nickel2016review}.

\begin{figure}[t]
\begin{minipage}[b]{0.95\linewidth}
  \centering
  \centerline{\includegraphics[width=\linewidth]{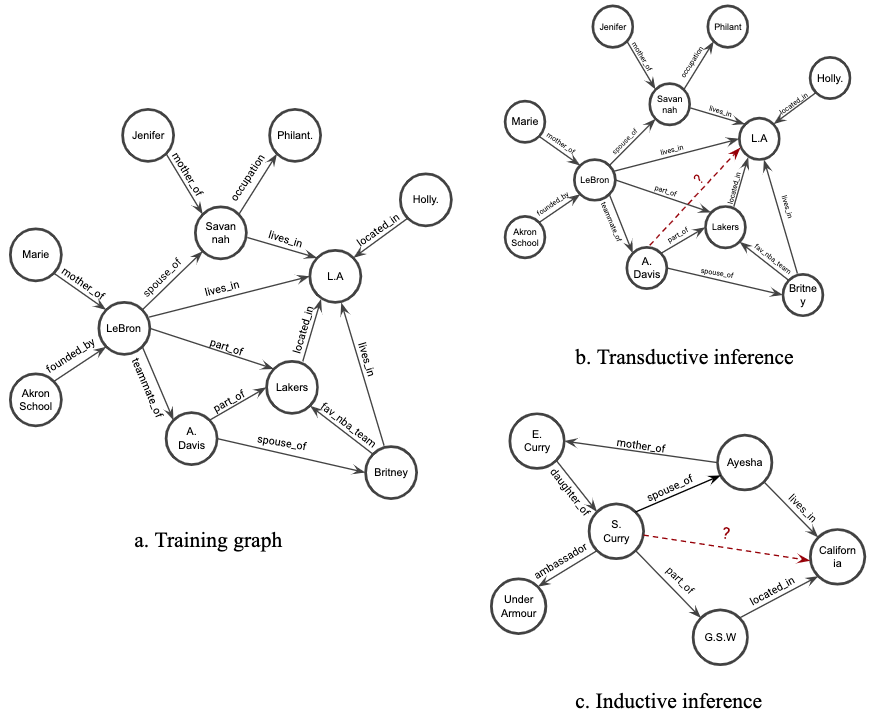}}
\end{minipage}
\caption{Illustration of transductive and inductive settings for relation prediction in knowledge graphs.}
\vspace{-10pt}
\label{fig:task}
\end{figure}

The most dominant paradigm, in recent times, has been to learn and operate on latent representations (i.e., embeddings) of entities and relations.
These methods condense each entity's neighborhood connectivity pattern into an entity-specific low-dimensional embedding, which can then be used to predict missing edges \cite{transE, complex, dettmers2018conve, sun2018rotate}.
For example, in Figure \@ \ref{fig:task}a, the embeddings of \texttt{LeBron} and \texttt{A.Davis} will contain the information that they are both part of the \texttt{Lakers} organization, which could later be retrieved to predict that they are teammates. 
Similarly, the pattern that anyone closely associated with the \texttt{Lakers} would live in \texttt{L.A} with high probability could be encoded in the embedding space.
Embedding-based methods have enjoyed great success by exploiting such local connectivity patterns and homophily. However, it is not clear if they effectively capture the \textit{relational semantics} of knowledge graphs---i.e., the logical rules that hold among the relations underlying the knowledge graph.

Indeed, the relation prediction task can also be viewed as a {\em logical induction} problem, where one seeks to derive probabilistic logical rules (horn clauses) underlying a given KG. For example, from the KG shown in Figure\@ \ref{fig:task}a one can derive the simple rule 
\begin{multline}\label{rule1}
    \exists Y. (X, \texttt{spouse\_of}, Y) \wedge (Y, \texttt{lives\_in}, Z)\\
    \rightarrow (X, \texttt{lives\_in}, Z).
\end{multline}
Using the example from Figure \ref{fig:task}b, this rule can predict the relation (\texttt{A.Davis}, \texttt{lives\_in}, \texttt{L.A}).
While the embedding-based methods encode entity-specific neighborhood information into an embedding, these logical rules capture entity-independent relational semantics.

\cut{
Current embedding-based methods are not suitable for the making predictions on entities that are unseen during training, as they ground themselves to the entities present in the training set and learn entity-specific embeddings.
}

One of the key advantages of learning  entity-independent relational semantics is the {\em inductive} ability to generalise to unseen entities.
For example, the rule in Equation \eqref{rule1} can naturally generalize to the unseen KG in Fig \ref{fig:task}c and predict the relation (\texttt{S.Curry}, \texttt{lives\_in}, \texttt{California}).

Whereas embedding-based approaches inherently assume a fixed set of entities in the graph---an assumption that is generally referred to as the \textit{transductive} setting (Figure\@ \ref{fig:task}) \cite{planetoid}---in many cases, we seek algorithms with the inductive capabilities afforded by inducing  entity-independent logical rules. 
Many real-world KGs are ever-evolving, with new nodes or entities being added over time---e.g., new users and products on e-commerce platforms or new molecules in biomedical knowledge graphs--- the ability to make predictions on such new entities without expensive re-training is essential for production-ready machine learning models.
Despite this crucial advantage of rule induction methods, they suffer from scalability issues and lack the expressive power of embedding-based approaches. 
\cut{The challenge of having to be able to generalize and make predictions on unseen nodes is referred to as the \textit{inductive} setting (Figure\@ \ref{fig:task}c) \cite{planetoid, hamilton2017inductive}. Using an inductive model, one could transfer knowledge from one domain to another. For example, a model trained on knowledge graph derived from one e-commerce platform could be used to make meaningful predictions on another e-commerce platform (with entirely different users and products) without having to re-train the model.}

\cut{
Current embedding-based methods are not suitable for the inductive setting, as they ground themselves to the entities present in the training set and learn entity-specific embeddings. For example, a embedding-model trained on KG in Fig \ref{fig:task}a can not make predictions on KG in Fig \ref{fig:task}c since it does not have an embedding for the new entities \texttt{s.curry} and \texttt{california}.
In contrast, the entity-independent logical rules can be used to make predictions on any set of entities irrespective of their presence during training.  Despite this crucial advantage of rule-based methods, they suffer from scalability issues and lack the expressive power that comes with distributed representations (i.e., the embeddings).}

% There are inductive approaches to generate embedding for unseen nodes (e.g., for node classification) \cite{hamilton2017inductive}. However, these approaches rely on the presence of node features which are not present in many KGs.
% Finally, in contrast to tasks such as node classification, relation prediction in an inductive setting is especially difficult since it requires the model to learn complex multi-hop rules that are independent of node identities. These rules need to be decoded solely from the structural/relational information in the graph, more so in the absence of any node features.

% \paragraph*{para 4 (key ideas/implications and contributions)}
% \begin{enumerate}
%     \item Key idea of the model (Instead of learning entity-specific embeddings, make predictions based on subgraph around the target link)
%     \item Things you can do with such a model (transfer knowledge from one domain to the other)
%     \item Motivation for dataset constructions (unavailability of benchmark inductive datasets)
%     \item Short description of baseline model (rule enumeration model).
%     \item Preview of results and conclusions.
% \end{enumerate}

\xhdr{Present work} 
We present a Graph Neural Network (GNN) \cite{scarselli2008graph, bronstein2017geometric}  framework (\name: \underline{Gra}ph \underline{I}nductive \underline{L}earning) that has a strong inductive bias to learn entity-independent relational semantics. 
In our approach, instead of learning entity-specific embeddings we learn to predict relations from the subgraph structure around a candidate relation. 
% We do not use any node attributes in order to test GNNs' ability to learn and generalize solely from structure.
% Since the GNN only receives structural information (i.e., the subgraph structure) as input, the only way it can complete the relation prediction task is to learn the structural semantic rules that underlie the knowledge graph.
We provide theoretical proof and strong empirical evidence that \name\ can represent logical rules of the kind presented above (e.g., Equation \eqref{rule1}).
Our approach naturally generalizes to unseen nodes, as the model learns to reason over subgraph structures independent of any particular node identities.

In addition to the \name\ framework, we also introduce a series of benchmark tasks for the inductive relation prediction problem.
Existing benchmark datasets for knowledge graph completion are set up for transductive reasoning, i.e., they ensure that all entities in the test set are present in the training data. 
Thus, in order to test models with inductive capabilities, we construct several new inductive benchmark datasets by carefully sampling subgraphs from diverse knowledge graph datasets. 
Extensive empirical comparisons on these novel benchmarks demonstrate that \name\ is able to substantially outperform state-of-the-art inductive baselines, with an average relative performance increase of $5.25\%$ and $6.75\%$ in AUC-PR and Hits@10, respectively, compared to the strongest inductive baseline.

Finally, we compare \name\ against existing embedding-based models in the transductive setting. In particular, we hypothesize that our approach has an inductive bias that is complementary to the embedding-based approaches, and we investigate the power of ensembling \name\ with embedding-based methods. 
We find that ensembling with \name\ leads to significant performance improvements in this setting.
\cut{
We perform pairwise ensembling of various methods and empirically show that the relative gains obtained by ensembling any embedding-based method with our approach is more than the gains obtained by ensembling any two embedding-based methods.
We also report the absolute gains obtained by performing early fusion of pre-trained embeddings, obtained from KGE methods, into our approach.
This highlights our framework's ability to naturally integrate any node features available.}

\cut{
we compare our model against the existing embedding-based models in the transductive setting. In particular, we hypothesize that our approach has an inductive bias that is complementary to the one in the embedding-based models.
We perform pairwise ensembling of various methods and empirically show that the relative gains obtained by ensembling any embedding-based method with our approach is more than the gains obtained by ensembling any two embedding-based methods.
We also report the absolute gains obtained by performing early fusion of pre-trained embeddings, obtained from KGE methods, into our approach.
This highlights our framework's ability to naturally integrate any node features available}
\cut{
Existing benchmark datasets for knowledge graph completion are set up for transductive reasoning, i.e., they ensure that all entities in the test set are present in the training data. 
Thus, in order to test models with inductive capabilities, we construct several new inductive benchmark datasets by carefully sampling subgraphs from diverse knowledge graph datasets. 
We evaluate our approach using these benchmarks against state-of-the-art rule-induction methods--both statistical and end-to-end differentiable neural architectures.
Our results show that the proposed GNN-based inductive model effectively generalizes to unseen nodes and graphs, outperforming the existing methods across all datasets.

In addition, we compare our model against the existing embedding-based models in the transductive setting. In particular, we hypothesize that our approach has an inductive bias that is complementary to the one in the embedding-based models.
We perform pairwise ensembling of various methods and empirically show that the relative gains obtained by ensembling any embedding-based method with our approach is more than the gains obtained by ensembling any two embedding-based methods.
We also report the absolute gains obtained by performing early fusion of pre-trained embeddings, obtained from KGE methods, into our approach.
This highlights our framework's ability to naturally integrate any node features available.}

% \newpage
\section{Related Work}
\label{sec:related}

\begin{figure*}
    \begin{minipage}[b]{\linewidth}
  \centering
  \centerline{\includegraphics[width=\linewidth]{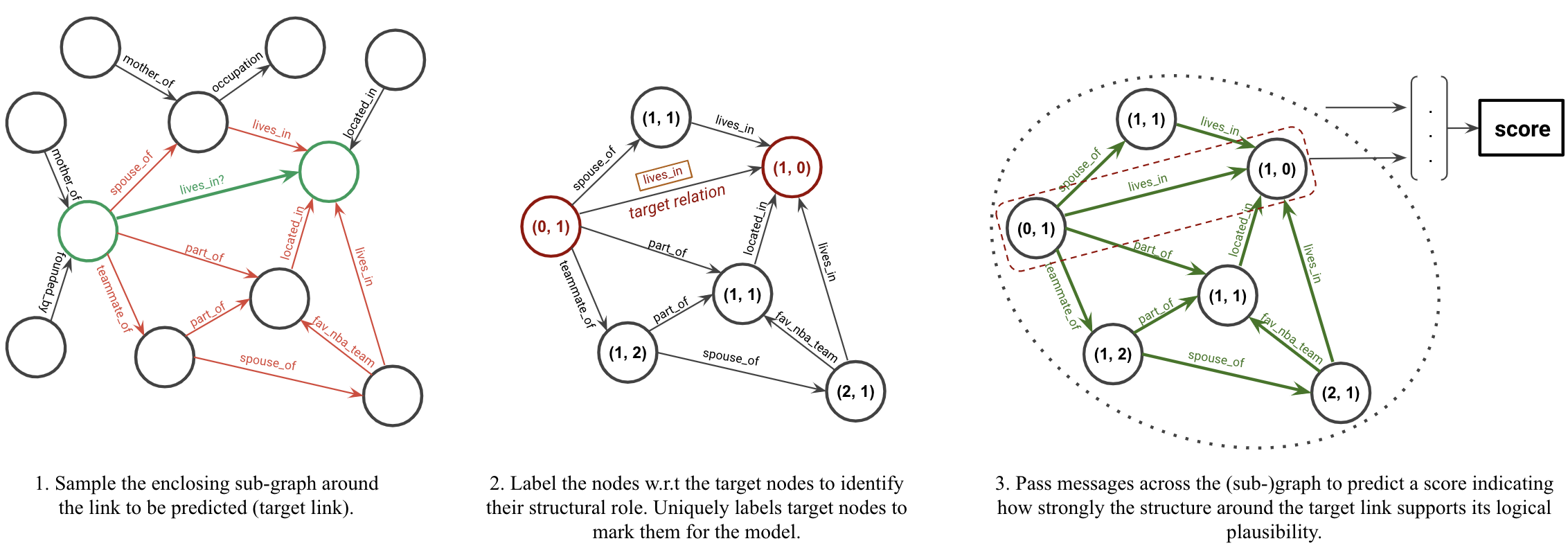}}
\end{minipage}
\caption{Visual illustration of \name\ for inductive relation prediction.}
\label{fig:modoverview}
\vspace{-10pt}
\end{figure*}

\xhdr{Embedding-based models} As noted earlier, most existing KG completion methods fall under the embedding-based paradigm. RotatE \cite{sun2018rotate}, ComplEx \cite{complex}, ConvE \cite{dettmers2018conve} and TransE \cite{transE} are some of the representative methods that train shallow embeddings \cite{Hamilton2017RepresentationLO} for each node in the training set, such that these low-dimensional embeddings can retrieve the relational information of the graph. Our approach embodies an alternative inductive bias to explicitly encode structural rules.
Moreover, while our framework is naturally inductive, adapting the embedding methods to make predictions in the inductive setting requires expensive re-training of embeddings for the new nodes.

Similar to our approach, the R-GCN model uses a GNN to perform relation prediction \cite{rgcn}. Although this approach, as originally proposed, is transductive in nature, it has the potential for inductive capabilities if given some node features \cite{hamilton2017inductive}.
Unlike our approach though, R-GCN still requires learning node-specific embeddings, whereas we treat relation prediction as a subgraph reasoning problem.

\xhdr{Inductive embeddings} There have been promising works for generating embeddings for unseen nodes, though they are limited in some ways. \citet{hamilton2017inductive} and \citet{Bojchevski2017DeepGE} rely on the presence of node features which are not present in many KGs. \cite{lan} and \cite{ookb} learn to generate embeddings for unseen nodes by aggregating neighboring node embeddings using GNNs. However, both of these approaches need the new nodes to be surrounded by known nodes and can not handle entirely new graphs.

%Our method might seem superficially similar to this method, but there are fundamental differences in the two approaches.
%While R-GCN learns the node-embeddings treating the graph as whole whole unit, we formulate the problem as directly predicting relation from subgraph structure around the target-link. This allows us to fully exploit the structural cues without necessarily depending on any node attributes or identities.

\xhdr{Rule-induction methods} Unlike embedding-based methods, statistical rule-mining approaches induce probabilistic logical-rules by enumerating statistical regularities and patterns present in the knowledge graph \cite{ruleN, amie}.
These methods are inherently inductive since the rules are independent of node identities, but these approaches suffer from scalability issues and lack expressive power due to their rule-based nature.
Motivated by these statistical rule-induction approaches, the NeuralLP model learns logical rules from KGs in an end-to-end differentiable manner \cite{neural-lp} using TensorLog \cite{tensorLog} operators. Building on NeuralLP, \citet{drum} recently proposed DRUM, which learns more accurate rules. This set of methods constitute our baselines in the inductive setting.

%This approach, in principle, is inductive. However, it requires the knowledge of all the entities the model will be tested on. Even though this information is not used to train the model, it is needed to initialize the TensorLog operators [TensorLog]. This ultimately limits its applicability to static graphs, unlike our approach which naturally scales to dynamic graphs.

\xhdr{Link prediction using GNNs}
Finally, outside of the KG literature, \citet{seal} have theoretically proven that GNNs can learn common graph heuristics for link prediction in simple graphs.
\citet{igmc} have used a similar approach to achieve competitive results on inductive matrix completion.
Our proposed approach can be interpreted as an extension of \citet{seal}s' method to directed multi-relational knowledge graphs.

% \todo{Briefly mention the link between SEAL/GraIL and Bruno's recent ICLR paper.}

\section{Proposed Approach}
\label{sec:approach}
% \todo{Inherent assumption that local neighborhood has enough information.}
% \todo{Details of attention and JK connection.}
% \paragraph*{para 1} : 
% \begin{enumerate}
%     \item High level idea of how our approach differs from existing methods.
% \end{enumerate}
    
The key idea behind our approach is to predict relation between two nodes from the subgraph structure around those two nodes. 
Our method is built around the Graph Neural Network (GNN) \cite{Hamilton2017RepresentationLO} (or Neural Message Passing \cite{mpnn}) formalism. 
% Let $\mathcal{G}(\mathcal{V}, \mathcal{E})$ denote the whole graph.
We do not use any node attributes in order to test \name's ability to learn and generalize solely from structure.
Since it only ever receives structural information (i.e., the subgraph structure and structural node features) as input, the only way \name\ can complete the relation prediction task is to learn the structural semantics that underlie the knowledge graph.
The overall task is to score a triplet $(u, r_t, v)$, i.e., to predict the likelihood of a possible relation $r_t$ between a \textit{head} node $u$ and \textit{tail} node $v$ in a KG, where we refer to nodes $u$ and $v$ as \textit{target nodes} and to $r_t$ as the \textit{target relation}. Our approach to scoring such triplets can be roughly divided into three sub-tasks (which we detail below): (i) extracting the enclosing subgraph around the target nodes, (ii) labeling the nodes in the extracted subgraph, and (iii) scoring the labeled subgraph using a GNN (Figure\@ \ref{fig:modoverview}). 

% We do so by training a graph neural network, on various subgraphs, to implicitly learn the compositional logical rules. Let $\mathcal{G}(\mathcal{V}, \mathcal{E})$ denote the whole graph. The task is to predict a score for the triplet $(u, r, v)$ i.e. score for a possible relation $r$ between the nodes $u$ and $v$ ($u \xrightarrow{r} v$). More generally, we would refer to nodes $u$ and $v$ as \textit{target} nodes and to $r$ as the \textit{target} relation.

% \paragraph*{para 2}:
% \begin{enumerate}
%     \item Describe how (enclosing) subgraphs are sampled around the target link.
%     \item Details: preserve direction, add the target link to allow message passing between target nodes.
% \end{enumerate}

\subsection{Model Details}\label{sec:model}
\xhdr{Step 1: subgraph extraction} 
% \todo{Emphasize the importance of \textbf{enclosing} subgraph over other obvious alternates.}
We assume that local graph neighborhood of a particular triplet in the KG will contain the logical evidence needed to deduce the relation between the target nodes.
In particular, we assume that the paths connecting the two target nodes contain the information that could imply the target relation.
Hence, as a first step, we extract the enclosing subgraph around the target nodes.
We define the \textit{enclosing subgraph} between nodes $u$ and $v$ as the graph induced by all the nodes that occur on a path between $u$ and $v$.
% In the experiments section, we demonstrate the importance of considering \textit{enclosing subgraphs}.
It is given by the intersection of neighbors of the two target nodes followed by a pruning procedure.
More precisely, let $\mathcal{N}_{k}(u)$ and $\mathcal{N}_{k}(v)$ be set of nodes in the $k$-hop (undirected) neighborhood of the two target nodes in the KG.
We compute the enclosing subgraph by taking the intersection,
$\mathcal{N}_{k}(u) \cap \mathcal{N}_{k}(v)$, of these $k$-hop neighborhood sets and then prune nodes that are isolated or at a distance greater than $k$ from either of the target nodes. 
Following the Observation 1, this would give us all the nodes that occur on a path of length at most $k + 1$ between nodes $u$ and $v$.

\xhdr{Observation 1} \textit{In any given graph, let the nodes on a path of length $\lambda$ between two different nodes $x$ and $y$ constitute the set $P_{xy}(\lambda)$. The maximum distance of any node on such a path, $v \in P_{xy}(\lambda)$, from either $x$ or $y$ is $\lambda - 1$.}

Note that while extracting the enclosing subgraph we ignore the direction of the edges. However, the direction is preserved while passing messages with Graph Neural Network, a point re-visited later. Also, the target tuple/edge $(u, r_t, v)$ is added to the extracted subgraph to enable message passing between the two target nodes.

% \paragraph*{para 3}
% \begin{enumerate}
%     \item Describe the previous practices followed for node-labeling -- one hot and node attributes/features -- and motivate why we use different node-labeling scheme.
%     \item Describe node-labeling process : do not forget target nodes labeling.
%     \item Describe how node features, if available, could naturally be integrated into the model.
% \end{enumerate}

\xhdr{Step 2: Node labeling}
GNNs require a node feature matrix, $\mathbf{X} \in \mathbb{R}^{|\mathcal{V}| \times d_i}$, as input, which is used to initialize the neural message passing algorithm \cite{mpnn}. 
Since we do not assume any node attributes in our input KGs, we follow \citet{seal} and extend their double radius vertex labeling scheme to our setting.
Each node, $i$, in the subgraph around nodes $u$ and $v$ is labeled with the tuple $(d(i, u), d(i, v))$, where $d(i, u)$ denotes the shortest distance between nodes $i$ and $u$ without counting any path through $v$ (likewise for $d(i, v)$). This captures the topological position of each node with respect to the target nodes and reflects its structural role in the subgraph. The two target nodes, $u$ and $v$, are uniquely labeled $(0, 1)$ and $(1, 0)$ so as to be identifiable by the model. The node features are thus $[\text{one-hot}(d(i, u)) \oplus \text{one-hot}(d(i, v))]$, where $\oplus$ denotes concatenation of two vectors. Note that as a consequence of Observation 1, the dimension of node features constructed this way is bounded by the number of hops considered while extracting the enclosing subgraph.

% \textbf{TODO: Add motivation for having unique labels for the target nodes. 
% TODO: Describe how node features, if available, could naturally be integrated into the model.}

% \paragraph*{para 4}
% \begin{enumerate}
%     \item Describe the GNN architecture with equations (leaving out attention, jk connections and edge-dropout in favour of brevity). Note that we add the target link to allow message passing between target nodes.
%     \item Describe the MLP input (and motivations) at the end of GNN to get prediction scores.
% \end{enumerate}

\xhdr{Step 3: GNN scoring} 
The final step in our framework is to use a GNN to score the likelihood of tuple $(u, r_t, v)$ given $\mathcal{G}_{(u, v, r_t)}$---the extracted and labeled subgraph around the target nodes. We adopt the general message-passing scheme described in \citet{Xu2018HowPA} where a node representation is iteratively updated by combining it with aggregation of it's neighbors' representation. In particular, the $k^{th}$ layer of our GNN is given by,
\begin{align}
    %\label{eq:neighbor_agg}
    \mathbf{a}_t^{k} &= \text{AGGREGATE}^{k} \left( \left\lbrace \mathbf{h}_s^{k-1}  : s \in \mathcal{N}(t) \right\rbrace, \mathbf{h}^{k-1}_t\right), \\
    %\label{eq:combine}
    \mathbf{h}_t^{k}   &= \text{COMBINE}^{k} \left( \mathbf{h}_t^{k-1}, \mathbf{a}_t^{k} \right),
\end{align}
where $\mathbf{a}_t^{k}$ is the aggregated message from the neighbors, $\mathbf{h}_t^{k}$ denotes the latent representation of node $t$ in the $k$-th layer, and $\mathcal{N}(t)$ denotes the set of immediate neighbors of node $t$. 
The initial latent node representation of any node $i$, $\mathbf{h}_i^0$, is initialized to the node features, $\mathbf{X}_i$, built according to the labeling scheme described in Step 2.
This framework gives the flexibility to plug in different AGGREGATE and COMBINE functions resulting in various GNN architectures.

Inspired by the multi-relational R-GCN \cite{rgcn} and edge attention, we define our AGGREGATE function as
$$
\mathbf{a}_t^{k}  = \sum_{r=1}^{R} \sum_{s \in \mathcal{N}_{r}(t)} \alpha_{rr_tst}^k \mathbf{W}_r^k \mathbf{h}_s^{k-1},
$$
where $R$ is the total number of relations present in the knowledge graph;  $\mathcal{N}_{r}(t)$ denotes the immediate outgoing neighbors of node $t$ under relation $r$; $\mathbf{W}_r^k$ is the transformation matrix used to propagate messages in the $k$-th layer over relation $r$; $\alpha_{rr_tst}^k$ is the edge attention weight at the $k$-th layer corresponding to the edge connecting nodes $s$ and $t$ via relation $r$. This attention weight, a function of the source node $t$, neighbor node $s$, edge type $r$ and the target relation to be predicted $r_t$, is given by
\begin{align*}
\mathbf{s} &= \text{ReLU}\left(\mathbf{A}_1^k[\mathbf{h}_s^{k-1} \oplus \mathbf{h}_t^{k-1} \oplus \mathbf{e}_r^{a} \oplus \mathbf{e}_{r_t} ^{a}] + \mathbf{b}_1^k\right) \\
\alpha_{rr_tst}^k &= \sigma\left(\mathbf{A}_2^k\mathbf{s} + \mathbf{b}_2^k\right).
\label{eq:attn}
\end{align*}
Here $\mathbf{h}^k_s$ and $\mathbf{h}^k_t$ denote the latent node representation of respective nodes at $k$-th layer of the GNN, $\mathbf{e}_r^{a}$ and $\mathbf{e}_{r_t}^{a}$ denote learned attention embeddings of respective relations.
Note that the attention weights are not normalized and instead come out of a sigmoid gate which regulates the information aggregated from each neighbor. 
As a regularization measure, we adopt the basis sharing mechanism, introduced by \cite{rgcn}, among the transformation matrices of each layer,  $\mathbf{W}_r^k$. We also implement a form of {\em edge dropout}, where edges are randomly dropped from the graph while aggregating information from the neighborhood.

The COMBINE function that yielded the best results is also derived from the R-GCN architecture. It is given by
\begin{equation}
\mathbf{h}_t^{k} = \text{ReLU}\left(\mathbf{W}_{self}^k \mathbf{h}_t^{k-1} + \mathbf{a}_t^{k}\right).
\end{equation}
With the GNN architecture as described above, we obtain the node representations after $L$ layers of message passing. 
A subgraph representation of $\mathcal{G}_{(u, v, r_t)}$  is obtained by average-pooling of all the latent node representations:
\begin{equation}
\mb{h}^L_{\mathcal{G}_{(u, v, r_t)}} = \frac{1}{|\mathcal{V}|} \sum_{i \in \mathcal{V}} \mb{h}_i^L,
\label{eq:gpool}
\end{equation}
where $\mathcal{V}$ denotes the set of vertices in graph $\mathcal{G}_{(u, v, r_t)}$.

Finally, to obtain the score for the likelihood of a triplet $(u, r_t, v)$, we concatenate four vectors---the subgraph representation ($\mb{h}^L_{\mathcal{G}_{(u, v, r_t)}}$), the target nodes' latent representations ($\mb{h}^L_u$ and $\mb{h}^L_v$), and a learned embedding of the target relation ($\mb{e}_{r_t} 
% \in \mathbb{R}^{d_e}
$)---and pass these concatenated representations through a linear layer:
\begin{equation}
    \text{score}(u, r_t, v) = \mathbf{W}^T [\mb{h}^L_{\mathcal{G}_{(u, v, r_t)}} \oplus \mb{h}^L_u \oplus \mb{h}^L_v \oplus \mb{e}_{r_t}].
\label{score}
\end{equation}
In our best performing model, in addition to using the node representations from the last layer, we also make use of representations from intermittent layers. 
This is inspired by the JK-connection mechanism introduced by \citet{jk}, which allows for flexible neighborhood ranges for each node. 
Addition of such JK-connections made our model's performance robust to the number of layers of the GNN. 
Precise implementation details of basis sharing, JK-connections and other model variants that were experimented with can be found in the Appendix.

\subsection{Training Regime}
\label{sec:training}

Following the standard and successful practice, we train the model to score positive triplets higher than the negative triplets using a noise-contrastive hinge loss \cite{transE}. More precisely, for each triplet present in the training graph, we sample a \textit{negative} triplet by replacing the head (or tail) of the triplet with a uniformly sampled random entity. 
We then use the following loss function to train our model via stochastic gradient descent:
\begin{equation}
    \mathcal{L} = \sum_{i = 1}^{|\mathcal{E}|} \text{max}(0, \text{score}(n_i) - \text{score}(p_i) + \gamma),
\end{equation}
where $\mathcal{E}$ is the set of all edges/triplets in the training graph; $p_i$ and $n_i$ denote the positive and negative triplets respectively; $\gamma$ is the margin hyperparameter.

\subsection{Theoretical Analysis}
We can show that the \name\ architecture is capable of encoding the same class of {\em path-based} logical rules that are used in popular rule induction models, such as RuleN \cite{ruleN} and NeuralLP \cite{neural-lp} and studied in recent work on logical reasoning using neural networks \cite{sinha2019clutrr}. 
For the sake of exposition, we equate edges $(u,r,v)$ in the knowledge graph with binary logical predicates $r(u,v)$ where an edge $(u,r,v)$ exists in the graph iff $r(u,v) = true$. 

\begin{theorem}\label{thm:main}
Let $\mathcal{R}$ be any logical rule (i.e., clause) on binary predicates of the form:
\begin{equation*}\label{eq:abs_rule}
    r_t(X,Y) \leftarrow r_1(X,Z_1) \land r_2(Z_1, Z_2) \land ... \land r_k(Z_{k-1},Y),
\end{equation*}
where $r_t,r_1,...,r_k$ are (not necessarily unique) relations in the knowledge graph, $X,Z_1,...,Z_k,Y$ are free variables that can be bound by arbitrary unique entities. %, and where we have that
%\begin{equation*}
%    r^*(X,Y) \leftarrow r_1(X,Z) \land r_2(Z,Y)
%\end{equation*}
%for the special case $k=2$.
For any such $\mathcal{R}$ there exists a parameter setting $\Theta$ for a \name\ model with $k$ GNN layers and where the dimension of all latent embeddings are $d=1$ such that 
$$\textrm{score}(u, r_t, v) \neq 0$$
%if and only if $r_t(x,y)$ is the head of rule $\mathcal{R}$ and
if and only if $\exists Z_1,...,Z_k$ where the body of $\mathcal{R}$ is satisfied with $X=u$ and $Y=v$.
\end{theorem}
Theorem \ref{thm:main} states that any logical rule corresponding to a path in the knowledge graph can be encoded by the model. \name\ will output a non-zero value if and only if the body of this logical rule evaluates to true when grounded on a particular set of query entities $X=u$ and $Y=v$.
The full proof of Theorem \ref{thm:main} is detailed in the Appendix, but the key idea is as follows: Using the edge attention weights it is possible to set the model parameters so that the hidden embedding for a node is non-zero after one round of message passing (i.e., $\mathbf{h}_s^i \neq 0$) if and only if the node $s$ has at least one neighbor by a relation $r_i$. 
In other words, the edge attention mechanism allows the model to indicate whether a particular relation is incident to a particular entity, and---since we have uniquely labeled the targets nodes $u$ and $v$---we can use this relation indicating property to detect the existence of a particular path between nodes $u$ and $v$.

We can extend Theorem \ref{thm:main} in a straightforward manner to also show the following:
\begin{corollary}\label{cor:count}
Let $\mathcal{R}_1...,\mathcal{R}_m$ be a set of logical rules with the same structure as in Theorem \ref{thm:main} where each rule has the same head $r_t(X,Y)$.
Let 
\begin{align*}
\beta = |\{\mathcal{R}_i \: : \: &\textrm{$\exists Z_1,...,Z_k$  where $\mathcal{R}_i = true$}\\ &\textrm{with $X=u$ and $Y=v$}\}|.
\end{align*}
Then there exists a parameter setting for \name\ with the same assumptions as Theorem \ref{thm:main} such that
\begin{equation*}
    \textrm{score}(u,r_t,v) \propto \beta.
\end{equation*}
\end{corollary}
This corollary shows that given a set of logical rules that implicate the same target relation, \name\ can count how many of these rules are satisfied for a particular set of query entities $u$ and $v$.
In other words, similar to rule-induction models such as RuleN, \name\ can combine evidence from multiple rules to make a prediction. 

Interestingly, Theorem \ref{thm:main} and Corollary \ref{cor:count} indicate that \name\ can learn logical rules using only one-dimensional embeddings of entities and relations, which dovetails with our experience that \name's performance is reasonably stable for dimensions in the range $d=1,...,64$. 
However, the above analysis only corresponds to a fixed class of logical rules, and we expect that \name\ can benefit from a larger latent dimensionality to learn different kinds of logical rules and more complex compositions of these rules.

\subsection{Inference Complexity}
\label{sec:limitations}

Unlike traditional embedding-based approaches, inference in the \name\ model requires extracting and processing a subgraph around a candidate edge $(u,r_t,v)$ and running a GNN on this extracted subgraph. 
Given that our processing requires evaluating shortest paths from the target nodes to all other nodes in the extracted subgraph, we have that the inference time complexity of \name\ to score a candidate edge $(u,r_t,v)$ is 
\begin{equation*}
O(\log(\mathcal{V})\mathcal{E} + \mathcal{R}dk),
\end{equation*}
where $\mathcal{V}$, $\mathcal{R}$, and $\mathcal{E}$ are the number of nodes, relations and edges, respectively, in the enclosing subgraph induced by $u$ and $v$. $d$ is the dimension of the node/relation embeddings.

Thus, the inference cost of \name\ depends largely on the size of the extracted subgraphs, and the runtime in practice is usually dominated by running Dijkstra’s algorithm on these subgraphs.

% \textbf{Concluding statement briefing the whole process in a line or two. Put up an illustrative figure. Perhaps an algorithm box.}

\section{Experiments}
\label{sec:exp}

\begin{table*}[t]
    \caption{Inductive results (AUC-PR)}
    \centering
    \begin{tabular}{@{}lllllllllllll@{}}
        \toprule
         & \multicolumn{4}{c}{WN18RR} & \multicolumn{4}{c}{FB15k-237} & \multicolumn{4}{c}{NELL-995}\\
        %  \cmidrule(lr){2-3} \cmidrule(lr){4-5} \cmidrule(lr){6-7} \\
        % \addlinespace[-12pt]
        & v1 & v2 & v3 & v4 & v1 & v2 & v3 & v4 & v1 & v2 & v3 & v4 \\
        % \addlinespace[10pt]
        % \cmidrule(lr){2-3} \cmidrule(lr){4-5} \cmidrule(lr){6-7} \\
        \midrule
        Neural-LP & 86.02 &	83.78 & 62.90 & 82.06 & 69.64 &	76.55 &	73.95 &	75.74 & 64.66 &	83.61 &	87.58 &	85.69\\
        DRUM & 86.02 &	84.05 & 63.20 & 82.06 & 69.71 &	76.44 &	74.03 &	76.20 & 59.86 &	83.99 &	\underline{87.71} &	\underline{85.94}\\
        RuleN & \underline{90.26} &	\underline{89.01} &	\underline{76.46} &	\underline{85.75} &	\underline{75.24} &	\underline{88.70} &	\underline{91.24} &	\underline{91.79} &	\underline{84.99} &	\underline{88.40} &	87.20 &	80.52 \\
        GraIL & \textbf{94.32} &\textbf{94.18} &	\textbf{85.80} &	\textbf{92.72} &	\textbf{84.69} &	\textbf{90.57} &	\textbf{91.68} &	\textbf{94.46} &	\textbf{86.05} &	\textbf{92.62} &	\textbf{93.34} &	\textbf{87.50} \\
        \bottomrule
    \end{tabular}
    \label{tab:ind_results_auc}
\end{table*}

\begin{table*}[t]
    \caption{Inductive results (Hits@10)}
    \centering
    \begin{tabular}{@{}lllllllllllll@{}}
        \toprule
         & \multicolumn{4}{c}{WN18RR} & \multicolumn{4}{c}{FB15k-237} & \multicolumn{4}{c}{NELL-995}\\
        %  \cmidrule(lr){2-3} \cmidrule(lr){4-5} \cmidrule(lr){6-7} \\
        % \addlinespace[-12pt]
        & v1 & v2 & v3 & v4 & v1 & v2 & v3 & v4 & v1 & v2 & v3 & v4 \\
        % \addlinespace[10pt]
        % \cmidrule(lr){2-3} \cmidrule(lr){4-5} \cmidrule(lr){6-7} \\
        \midrule
        Neural-LP & 74.37 &	68.93 &	46.18 &	67.13 & \underline{52.92} &	58.94 &	52.90 &	55.88 & 40.78 &	78.73 &	\underline{82.71} &	\textbf{80.58}\\
        DRUM & 74.37 &	68.93 & 46.18 & 67.13 & \underline{52.92} &	58.73 &	52.90 &	55.88 & 19.42 &	78.55 &	\underline{82.71} &	\textbf{80.58}\\
        RuleN & \underline{80.85} &	\underline{78.23} &	\underline{53.39} &	\underline{71.59} & 49.76 &	\underline{77.82} &	\textbf{87.69} &	\underline{85.60} & \underline{53.50} &	\underline{81.75} &	77.26 &	61.35\\
        GraIL & \textbf{82.45} &	\textbf{78.68} &	\textbf{58.43} &	\textbf{73.41} & \textbf{64.15} &	\textbf{81.80} &	\underline{82.83} &	\textbf{89.29} & \textbf{59.50} &	\textbf{93.25} &	\textbf{91.41} &	\underline{73.19}\\
        \bottomrule
    \end{tabular}
    \label{tab:ind_results_hits}
\end{table*}

We perform experiments on three benchmark knowledge completion datasets: WN18RR \cite{dettmers2018conve}, FB15k-237 \cite{toutanova2015representing}, and NELL-995 \cite{deeppath} (and other variants derived from them). Our empirical study is motivated by the following questions:
\begin{enumerate}[itemsep=3pt,topsep=1pt]
    \item \xhdr{Inductive relation prediction} By Theorem 1, we know that \name\ can encode inductive logical rules. How does it perform in comparison to existing statistical and differentiable methods which explicitly do rule induction in the inductive setting?
    
    \item \xhdr{Transductive relation prediction} Our approach has a strong structural inductive bias which, we hypothesize, is complementary to existing state-of-the-art knowledge graph embedding methods. Can this complementary inductive bias give any improvements over the existing state-of-the-art KGE methods in the traditional transductive setting?
    
    \item \xhdr{Ablation study} How important are the various components of our proposed framework? For example, Theorem \ref{thm:main} relies on the use of attention and the node-labeling scheme, but how important are these model aspects in practice?
\end{enumerate}
The code and the data for all the following experiments is available at: \small{\url{https://github.com/kkteru/grail}}.
\normalsize
\cut{
We study the relation prediction abilities of \name\ in both \textit{inductive} and \textit{transductive} settings on three different benchmark knowledge graphs: WN18RR \cite{dettmers2018conve}, FB15k-237 \cite{toutanova2015representing}, and NELL-995 \cite{deeppath}.
WN18RR and FB15k-237 are refined subsets of WordNet and FreeBase with inverse relations removed to avoid any leakage into the test set. NELL-995, as presented by \cite{deeppath}, is a subset of 995th iteration of the knowledge graph collected by NELL system \cite{nell}. 
We remove certain relations from NELL-995 (e.g., latitude and longitude information of places), which do not contribute to the logical deduction of missing relations.
}

\subsection{Inductive Relation Prediction}\label{sec:ind_results}

As illustrated in Figure\@ \ref{fig:task}c, an inductive setting evaluates a models' ability to generalize to unseen entities. In a fully inductive setting the sets of entities seen during training and testing are disjoint. More generally, the number of unseen entities can be varied from only a few new entities being introduced to a fully-inductive setting (Figure \ref{fig:task}c). 
The proposed framework, \name, is invariant to the node identities so long as the underlying semantics of the relations (i.e., the schema of the knowledge graph) remains the same. We demonstrate our inductive results in the extreme case of having an entirely new test graph with new set of entities.

\xhdr{Datasets} The WN18RR, FB15k-237, and NELL-995  benchmark datasets  were originally developed for the transductive setting. 
In other words, the entities of the standard test splits are a subset of the entities in the training splits (Figure\ref{fig:task}b).
In order to facilitate inductive testing, we create new fully-inductive benchmark datasets by sampling disjoint subgraphs from the KGs in these datasets. In particular, each of our datasets consist of a pair of graphs: \textit{train-graph} and \textit{ind-test-graph}. These two graphs (i) have disjoint set of entities and (ii) \textit{train-graph} contains all the relations present in \textit{ind-test-graph}. The procedure followed to generate such pairs is detailed in the Appendix. For robust evaluation, we sample four different pairs of \textit{train-graph} and \textit{ind-test-graph} with increasing number of nodes and edges for each benchmark knowledge graph. The statistics of these inductive benchmarks is given in Table \ref{table:data-stats} in the Appendix.
In the inductive setting, a model is trained on \textit{train-graph} and tested on \textit{ind-test-graph}. We randomly select 10\% of the edges/tuples in \textit{ind-test-graph} as test edges. 

\xhdr{Baselines and implementation details} 
We compare \name\ with two other end-to-end differentiable methods, NeuralLP \cite{neural-lp} and DRUM \cite{drum}. To the best of our knowledge, these are the only differentiable methods capable of inductive relation prediction. We use the implementations publicly provided by the authors with their best configurations.
We also compare against a state-of-the-art statistical rule mining method, RuleN \cite{ruleN}, which performs competitively with embedding-based methods in the transductive setting. 
RuleN represents the current state-of-the-art in inductive relation prediction on KGs.
It explicitly extracts path-based rules of the kind as shown in Equation \eqref{rule1}. 
Using the original terminology of RuleN, we train it to learn rules of length up to 4. By Observation 1 this corresponds to 3-hop neighborhoods around the target nodes.
In order to maintain a fair comparison, we sample 3-hop enclosing subgraphs around the target links for our GNN approach. We employ a 3-layer GNN with the dimension of all latent embeddings equal to 32.
The basis dimension is set to 4 and the edge dropout rate to $0.5$.
In our experiments, \name\ was relatively robust to hyperparameters and had a stable performance across a wide range of settings.
Further hyperparameter choices are detailed in the Appendix.

\cut{
Our primary baseline is RuleN \cite{ruleN} ---a recently proposed rule-based method, which performs competitively with embedding-based methods in the transductive setting and outperforms other rule-based approaches (e.g., AMIE \cite{amie}). 
RuleN represents the current state-of-the-art in inductive relation prediction on KGs. 
It explicitly extracts horn clause rules of the kind as shown in Equation \eqref{rule1}. 
Using the original terminology of RuleN, we train it to learn rules of length up to 4. According to Observation 1 this corresponds to 3-hop neighborhood of target nodes.
For a fair comparison, we sample 3-hop enclosing subgraphs around the target links for our GNN approach. We employ a 3-layer GNN and the embedding size of each hidden layer (as well as the relation embeddings: $\mathbf{e}_r$ and $\mathbf{e}^a_r$) is 32.
For the basis sharing, we set the basis dimension to be fixed at 4. We fix the edge dropout rate to $0.5$.
Further hyperparameter choices are detailed in the appendix.
In addition to RuleN we also compare our model with Neural-LP \cite{neural-lp}, which is an end-to-end differentiable rule-mining method.
}

\xhdr{Results} We evaluate the models on both classification and ranking metrics, i.e., area under precision-recall curve (AUC-PR) and Hits@10 respectively. 
To calculate the AUC-PR, along with the triplets present in the test set, we score an equal number of negative triplets sampled using the standard practice of replacing head (or tail) with a random entity.
To evaluate Hits@10, we rank each test triplet among 50 other randomly sampled negative triplets.
In Table \ref{tab:ind_results_auc} and Table \ref{tab:ind_results_hits} we report the mean AUC-PR and Hits@10, respectively, averaged over 5 runs. (The variance was very low in all the settings, so the standard errors are omitted in these tables.)

As we can see, \name\ significantly outperforms the inductive baselines across all datasets in both metrics.
This indicates that \name\ is not only able to learn path-based logical rules, which are also learned by RuleN, but that \name\ is able to also exploit more complex structural patterns.
For completeness, we also report the transductive performance on these generated datasets in the Appendix. 
Note that the inductive performance (across all datasets and models) is relatively lower than the transductive performance, highlighting the difficulty of the inductive relation prediction task.

\subsection{Transductive Relation Prediction}\label{sec:trans_res}

As demonstrated, \name\ has a strong inductive bias to encode the logical rules and complex structural patterns underlying the knowledge graph. 
This, we believe, is complementary to the current state-of-the-art transductive methods for knowledge graph completion, which rely on embedding-based approaches. 
Based on this observation, in this section we explore (i) how \name\ performs in the transductive setting and (ii) the utility of ensembling \name\ with existing embedding-based approaches.
Given \name's complementary inductive bias compared to embedding-based methods, we expect significant gains to be obtained by ensembling it with existing embedding-based approaches. 
%Ensemble of such diverse and accurate methods generally results in better performance \cite{hansen1990neural}.
%Hence, we ensemble \name\ with different transductive methods and note the relative gains obtained.

Our primary ensembling strategy is late fusion i.e., ensembling the output scores of the constituent methods. We score each test triplet with the methods that are to be ensembled. The scores output by each method form the feature vector for each test point. This feature vector is input to a linear classifier which is trained to score the true triplets higher than the negative triplets. We train this linear classifier using the validation set. 
%At test time, the test set is scored by the constituent methods generating the feature vector which is then input to the linear classifier for the final score.

\cut{
Early fusion refers to using the embeddings learnt by transductive methods as additional input into \name. In particular, the node labels, as computed by our original node-labeling scheme, are concatenated with node-embeddings learnt by a transductive method. This would test \name's ability to exploit auxiliary node information, when available.
}

\xhdr{Datasets} We use the standard WN18RR, FB15k-237, and NELL-995 benchmarks. For WN18RR and FB15k-237, we use the splits as available in the literature. For NELL-995, we split the whole dataset into train/valid/test set by the ratio 70/15/15, making sure all the entities and relations in the valid and test splits occur at least once in the train set.

\xhdr{Baselines and implementation details} We ensemble \name\ with each of TransE \cite{transE}, DistMult \cite{distmult}, ComplEx \cite{complex}, and RotatE \cite{sun2018rotate} which constitute a representative set of KGE methods. For all the methods we use the implementation and hyperparameters provided by \citet{sun2018rotate} which gives state-of-the-art results on all methods. For a fair comparison of all the methods, we disable the self-adversarial negative sampling proposed by \citet{sun2018rotate}. For \name, we use 2-hop neighborhood subgraphs for WN18RR and NELL-995, and 1-hop neighborhood subgraphs for FB15k-237. All the other hyperparameters for \name\ remain the same as in the inductive setting.

\begin{table}[t]
    \caption{Late fusion ensemble results on WN18RR (AUC-PR)}
    \centering
    \begin{tabular}{@{}llllll@{}}
        \toprule
        & TransE & DistMult & ComplEx & RotatE & GraIL \\
        \midrule
        %  \cmidrule(lr){2-3} \cmidrule(lr){4-5} \cmidrule(lr){6-7} \\
        % \addlinespace[-12pt]
        T & 93.73 &	93.12 &	92.45 &	93.70 &	94.30 \\
        D & & 93.08 &	93.12 &	93.16 &	95.04 \\
        C & & & 92.45 & 92.46 &	94.78 \\
        R & & & & 93.55 &	94.28 \\
        G & & & & & 90.91 \\
        % \addlinespace[10pt]
        % \cmidrule(lr){2-3} \cmidrule(lr){4-5} \cmidrule(lr){6-7} \\
        \bottomrule
    \end{tabular}
    \label{tab:ens_late_wn}
\end{table}

\begin{table}[t]
    \caption{Late fusion ensemble results on NELL-995 (AUC-PR)}
    \centering
    \begin{tabular}{@{}llllll@{}}
        \toprule
        & TransE & DistMult & ComplEx & RotatE & GraIL \\
        \midrule
        %  \cmidrule(lr){2-3} \cmidrule(lr){4-5} \cmidrule(lr){6-7} \\
        % \addlinespace[-12pt]
        T & 98.73 &	98.77 &	98.83 &	98.71 &	98.87 \\
        D & & 97.73 &	97.86 &	98.60 &	98.79 \\
        C & & & 97.66 &	98.66 &	98.85 \\
        R & & & & 98.54 &	98.75 \\
        G & & & & & 97.79 \\
        % \addlinespace[10pt]
        % \cmidrule(lr){2-3} \cmidrule(lr){4-5} \cmidrule(lr){6-7} \\
        \bottomrule
    \end{tabular}
    \label{tab:ens_late_nell}
\end{table}

\begin{table}[t]
    \caption{Late fusion ensemble results on FB15k-237 (AUC-PR)}
    \centering
    \begin{tabular}{@{}llllll@{}}
        \toprule
        & TransE & DistMult & ComplEx & RotatE & GraIL \\
        \midrule
        %  \cmidrule(lr){2-3} \cmidrule(lr){4-5} \cmidrule(lr){6-7} \\
        % \addlinespace[-12pt]
        T & 98.54 &	98.41 &	98.45 &	98.55 &	97.95 \\
        D & & 97.63 &	97.87 &	98.40 &	97.45 \\
        C & & & 97.99 &	98.43 &	97.72 \\
        R & & & & 98.53 &	98.04 \\
        G & & & & & 92.06 \\
        % \addlinespace[10pt]
        % \cmidrule(lr){2-3} \cmidrule(lr){4-5} \cmidrule(lr){6-7} \\
        \bottomrule
    \end{tabular}
    \label{tab:ens_late_fb}
\end{table}

% \begin{table}[t]
%     \caption{Relative gain of pairwise ensembling AUC-PR}
%     \centering
%     \begin{tabular}{@{}llll@{}}
%         \toprule
%          & WN18RR & FB15k-237 & NELL-995 \\
% 	\midrule
%         %  \cmidrule(lr){2-3} \cmidrule(lr){4-5} \cmidrule(lr){6-7} \\
%         % \addlinespace[-12pt]
% 	$G_{\text{avg}}^{\text{\name}}$ & 1.5\% & 0\% & 0.62\% \\
% 	$G_{\text{avg}}^{\text{KGE}}$ & 0.007\% & 0.002\% & 0.08\% \\
%         \bottomrule
%     \end{tabular}
%     \label{tab:rel_gains}
% \end{table}

\xhdr{Results} Tables \ref{tab:ens_late_wn}, \ref{tab:ens_late_fb}, and \ref{tab:ens_late_nell} show the AUC-PR performance of pairwise ensembling of different KGE methods among themselves and with \name.
A specific entry in these tables corresponds to the ensemble of pair of methods denoted by the row and column labels, with the individual performance of each method on the diagonal. 
As can be seen from the last column of these tables, ensembling with \name\ resulted in consistent performance gains across all transductive methods in two out of the three datasets. Moreover, ensembling with \name\ resulted in more gains than ensembling any other two methods. Precisely, we define the gain obtained by ensembling two methods, $G(M_1, M_2)$, as follows
$$
G(M_1, M_2) = \frac{P(M_1, M_2) - max(P(M_1), P(M_2))}{max(P(M_1), P(M_2)}.
$$

% \begin{table*}[h]
%     \caption{Early fusion ensemble with TrasnE results}
%     \centering
%     \begin{tabular}{@{}lllllll@{}}
%         \toprule
%          & \multicolumn{2}{c}{WN18RR} & \multicolumn{2}{c}{FB15k-237} & \multicolumn{2}{c}{NELL-995}\\
%         %  \cmidrule(lr){2-3} \cmidrule(lr){4-5} \cmidrule(lr){6-7} \\
%         % \addlinespace[-12pt]
%         & AUC-PR & Hits@10 & AUC-PR & Hits@10 & AUC-PR & Hits@10 \\
%         % \addlinespace[10pt]
%         % \cmidrule(lr){2-3} \cmidrule(lr){4-5} \cmidrule(lr){6-7} \\
%         \midrule
%         GraIL & 90.91 & 73.12 &	92.06 & 75.87 & 97.79 & 94.54 \\
%         GraIL++ & 96.20 & 88.59 & 93.91 & 89.68 & 98.11 & 97.93 \\
%         \bottomrule
%     \end{tabular}
%     \label{tab:ens_early}
% \end{table*}

\begin{table}[t]
    \caption{Early fusion ensemble with TrasnE results (AUC-PR)}
    \centering
    \begin{tabular}{@{}llll@{}}
        \toprule
         & WN18RR & FB15k-237 & NELL-995\\
        %  \cmidrule(lr){2-3} \cmidrule(lr){4-5} \cmidrule(lr){6-7} \\
        % \addlinespace[-12pt]
        % & AUC-PR & Hits@10 & AUC-PR & Hits@10 & AUC-PR & Hits@10 \\
        % \addlinespace[10pt]
        % \cmidrule(lr){2-3} \cmidrule(lr){4-5} \cmidrule(lr){6-7} \\
        \midrule
        GraIL & 90.91 &	92.06 & 97.79 \\
        GraIL++ & 96.20 & 93.91 & 98.11 \\
        \bottomrule
    \end{tabular}
    \label{tab:ens_early}
\end{table}

In other words, it is the percentage improvement achieved relative to the best of the two methods. Thus, the average gain obtained by ensembling with \name\ is given by 
$$
G_{\text{avg}}^{\text{\name}} = \frac{1}{4} \sum_{|M_1| \in KGE} G(M_1, \text{\name}),
$$
and the average gain obtained by pairwise ensembling among the KGE methods is given by,
$$
G_{\text{avg}}^{\text{KGE}} = \frac{1}{12} \sum_{(|M_1|, |M_2|) \in KGE} G(M_1, M_2).
$$

The average gain obtained by \name\ on WN18RR and NELL-995 are $1.5\%$ and $0.62\%$, respectively. This is orders of magnitude better than the average gains obtained by KGE ensembling: $0.007\%$ and $0.08\%$. Surprisingly, none of the ensemblings resulted in significant gains on FB15k-237.
Thus, while \name\ on its own is optimized for the inductive setting and not state-of-the-art for transductive prediction, it does give a meaningful improvement over state-of-the-art transductive methods via ensembling. 

\cut{
The relative gains on all datasets are summarized in Table \ref{tab:rel_gains}. \name\ gives clearly better gains on WN18RR and NELL-995 datasets. On FB15k-237, while \name\ doesn't give any gains, ensemble of other methods don't give statistically significant gains either.
}

On a tangential note, Table \ref{tab:ens_early} shows the performance of \name\ when the node features, as computed by our original node-labeling scheme, are concatenated with node-embeddings learnt by a TransE model. 
The addition of these pre-trained embeddings results in as much as $18.2\%$ performance boost.
Thus, while \textit{late fusion} demonstrates the complementary inductive bias that \name\ embodies, this kind of \textit{early fusion} demonstrates the natural ability of \name\ to leverage any node embeddings/features available. All the Hits@10 results which display similar trends are reported in the Appendix.

\subsection{Ablation Study}

In this section, we emphasize the importance of the three key components of \name: i) enclosing subgraph extraction ii) double radius node labeling scheme, and iii) attention in the GNN. The results are summarized in Table \ref{tab:ablation}.
\begin{table}[t]
    \caption{Ablation study of the proposed framework (AUC-PR)}
    \centering
    \begin{tabular}{@{}lll@{}}
        \toprule
         & FB (v3) & NELL (v3) \\
	\midrule
        %  \cmidrule(lr){2-3} \cmidrule(lr){4-5} \cmidrule(lr){6-7} \\
        % \addlinespace[-12pt]
	\name\ & \textbf{91.68} & \textbf{93.34} \\
	\name\ w/o enclosing subgraph & 84.25 & 85.89 \\
	\name\ w/o node labeling scheme & 82.07 & 84.46 \\
	\name\ w/o attention in GNN & 90.27 & 87.30 \\
        \bottomrule
    \end{tabular}
    \label{tab:ablation}
\end{table}

\xhdr{Enclosing subgraph extraction} As mentioned earlier, we assume that the logical evidence for a particular link can be found in the subgraph surrounding the two target nodes of the link. Thus we proposed to extract the subgraph induced by all the nodes occurring on a path between the two target nodes. Here, we want to emphasize the importance of extracting only the paths as opposed to a more naive choice of extracting the subgraph induced by all the $k$-hop neighbors of the target nodes. The performance drastically drops in such a configuration. In fact, the model catastrophically overfits to the training data with training AUC of over 99\%. This pattern holds across all the datasets.

\xhdr{Double radius node labeling} Proof of Theorem \ref{thm:main} assumes having uniquely labeled target nodes, $u$ and $v$. We highlight the importance of this by evaluating \name\ with constant node labels of $(1, 1)$ instead of the originally proposed node labeling scheme. The drop in performance emphasizes the importance of our node-labeling scheme.%in helping \name\ find the logical pathts and reinforcing the topology of the subgraph.

\xhdr{Attention in the GNN} As noted in the proof of Theorem \ref{thm:main}, the attention mechanism is a vital component of our model in encoding the path rules. We evaluate \name\ without the attention mechanism and note significant performance drop, which echos with our theoretical findings.

% \begin{table}[t]
%     \caption{Ablation study on subset of FB15k-237 (AUC-PR)}
%     \centering
%     \begin{tabular}{@{}lll@{}}
%         \toprule
%          & Transductive & Inductive\\
%         % \addlinespace[10pt]
%         % \cmidrule(lr){2-3} \cmidrule(lr){4-5} \cmidrule(lr){6-7} \\
%         \midrule
%         Basic GNN & 92.14 & 88.31 \\
%         + JK connections & 93.68 & 89.87 \\
%         + attention + edge dropout & 95.36 & 90.18 \\
%         \bottomrule
%     \end{tabular}
%     \label{tab:ablation}
% \end{table}

\section{Conclusion}
\label{sec:conclusion}

We proposed a GNN-based framework, \name, for inductive knowledge graph reasoning.
Unlike embedding-based approaches, \name\ model is able to predict relations between nodes that were unseen during training and achieves state-of-the-art results in this inductive setting. 
Moreover, we showed that \name\ brings an inductive bias complementary to the current state-of-the-art knowledge graph completion methods. In particular, we demonstrated, with a thorough set of experiments, performance boosts to various knowledge graph embedding methods when ensembled with \name. In addition to these empirical results, we provide theoretical insights into the expressive power of GNNs in encoding a useful subset of logical rules.
%\todo{Can we say anything about implications of this theoretical insight in a broader sense, say, for other domains/applications?}

This work---with its comprehensive study of existing methods for inductive relation prediction and a set of new benchmark datasets---opens a new direction for exploration on inductive reasoning in the context of knowledge graphs. 
For example, obvious directions for further exploration include  extracting interpretable rules and structural patterns from \name,  analyzing how shifts in relation distributions impact inductive performance, and combining \name\ with meta learning strategies to handle the few-shot learning setting. 

\paragraph{Acknowledgements}
This research was funded in part by an academic grant from Microsoft Research, as well as a Canada CIFAR Chair in AI, held by Prof. Hamilton. Additionally, IVADO provided support to Etienne through the Undergraduate Research Scholarship.
\bibliography{main_arxiv}
\bibliographystyle{icml2020_arxiv}

%%%%%%%%%%%%%%%%%%%%%%%%%%%%%%%%%%%%%%%%%%%%%%%%%%%%%%%%%%%%%%%%%%%%%%%%%%%%%%%
%%%%%%%%%%%%%%%%%%%%%%%%%%%%%%%%%%%%%%%%%%%%%%%%%%%%%%%%%%%%%%%%%%%%%%%%%%%%%%%

\clearpage
\appendix

\section{JK Connections}
As mentioned earlier, our best performing model uses a JK-Connections in the scoring function, as given by, 
% layer. With the JK-Connection Equation \eqref{score}
\begin{equation}
    \text{score}(u, r_t, v) = \mathbf{W}^T \bigoplus_{i=1}^L[h^i_{\mathcal{G}_{(u, v, r_t)}} \oplus h^i_u \oplus h^i_v \oplus e_{r_t}].
\label{score_jk}
\end{equation}
This is inspired by \cite{jk} which lets the model adapt the effective neighborhood size for each node as needed. Empirically, this made our model's performance more robust to other number of GNN layers.

\begin{table*}[b]
    \caption{Transductive results (AUC-PR)}
    \centering
    \begin{tabular}{@{}lllllllllllll@{}}
        \toprule
         & \multicolumn{4}{c}{WN18RR} & \multicolumn{4}{c}{FB15k-237} & \multicolumn{4}{c}{NELL-995}\\
        %  \cmidrule(lr){2-3} \cmidrule(lr){4-5} \cmidrule(lr){6-7} \\
        % \addlinespace[-12pt]
        & v1 & v2 & v3 & v4 & v1 & v2 & v3 & v4 & v1 & v2 & v3 & v4 \\
        % \addlinespace[10pt]
        % \cmidrule(lr){2-3} \cmidrule(lr){4-5} \cmidrule(lr){6-7} \\
        \midrule
        RuleN & 81.79          & 83.97          & 81.51          & 82.63          & 87.07          & 92.49          & 94.26          & 95.18          & 80.16          & 87.87          & 86.89          & 84.45          \\
GraIL & \textbf{89.00} & \textbf{90.66} & \textbf{88.61} & \textbf{90.11} & \textbf{88.97} & \textbf{93.78} & \textbf{95.04} & \textbf{95.68} & \textbf{83.95} & \textbf{92.73} & \textbf{92.30} & \textbf{89.29}\\
        \bottomrule
    \end{tabular}
    \label{tab:trans_results_auc}
\end{table*}

\begin{table*}[b]
    \caption{Transductive results (Hits@10)}
    \centering
    \begin{tabular}{@{}lllllllllllll@{}}
        \toprule
         & \multicolumn{4}{c}{WN18RR} & \multicolumn{4}{c}{FB15k-237} & \multicolumn{4}{c}{NELL-995}\\
        %  \cmidrule(lr){2-3} \cmidrule(lr){4-5} \cmidrule(lr){6-7} \\
        % \addlinespace[-12pt]
        & v1 & v2 & v3 & v4 & v1 & v2 & v3 & v4 & v1 & v2 & v3 & v4 \\
        % \addlinespace[10pt]
        % \cmidrule(lr){2-3} \cmidrule(lr){4-5} \cmidrule(lr){6-7} \\
        \midrule
        RuleN & 63.42          & 68.09          & 63.05          & 65.55          & 67.53          & \textbf{88.00} & \textbf{91.47} & \textbf{92.35} & 62.82          & 82.82          & 80.72          & 58.84          \\
GraIL & \textbf{65.59} & \textbf{69.36} & \textbf{64.63} & \textbf{67.28} & \textbf{71.93} & 86.30          & 88.95          & 91.55          & \textbf{64.08} & \textbf{86.88} & \textbf{84.19} & \textbf{82.33}\\
        \bottomrule
    \end{tabular}
    \label{tab:trans_results_hits}
\end{table*}

\section{Other Model Variants}
As mentioned in Section \ref{sec:model}, the formulation of our GNN scoring model allows for flexibility to plug in different AGGREGATE and COMBINE functions. We experimented with pooling AGGRAGATE function \citet{hamilton2017inductive} and two other COMBINE function (similar to CONCAT operation from \citet{hamilton2017inductive} and using a GRU as in \citet{li2015gated}). None of these variants gave significant improvements in the performance.

\section{Hyperparameter Settings}
The model was implemented in PyTorch. Experiments were run for 50 epochs on a GTX 1080 Ti with 12 GB RAM. The Adam optimizer was used with a learning rate of 0.01, L2 penalty of 5e-4, and default values for other parameters. 
The margin in the loss was set to 10. 
Gradient were clipped at a norm of 1000. The model was evaluated on the validation and saved every three epochs with the best performing checkpoint used for testing. 

\section{Transductive Results}
The transductive results, as mentioned in the discussion on inductive results (Section \ref{sec:ind_results}), were obtained using the the same methodology of the aforementioned evaluations. In particular, \name\ was trained on the \textit{train-graph} and tested n the same. We randomly selected $10\%$ of the links in \textit{train-graph} as test links. Tables \ref{tab:trans_results_auc} and \ref{tab:trans_results_hits} showcase the transductive results. The AUC-PR and Hits@10 in the transductive setting are significantly better than in the inductive setting, establishing the difficulty of the inductive task. \name\ performs significantly better than RuleN in most cases and is competitive in others.

\section{Inductive Setting Hits@10}
The Hits@10 results for the late fusion models in the transductive setting complementing the tables \ref{tab:ens_late_wn}, \ref{tab:ens_late_fb} and \ref{tab:ens_late_nell}  are given below. Similar trends, as discussed in Section \ref{sec:trans_res}, hold here as well.
\\
\begin{table}[h]
    \caption{Late fusion ensemble results on WN18RR (Hits@10)}
    \centering
    \begin{tabular}{@{}llllll@{}}
        \toprule
        & TransE & DistMult & ComplEx & RotatE & GraIL \\
        \midrule
        %  \cmidrule(lr){2-3} \cmidrule(lr){4-5} \cmidrule(lr){6-7} \\
        % \addlinespace[-12pt]
        T & 88.74 &	85.31 &	83.84 &	88.61 &	89.71 \\
        D & & 85.35 &	86.07 &	85.64 &	87.70 \\
        C & & & 83.98 &	84.30 &	86.73 \\
        R & & & & 88.85 &	89.84 \\
        G & & & & & 73.12 \\
        % \addlinespace[10pt]
        % \cmidrule(lr){2-3} \cmidrule(lr){4-5} \cmidrule(lr){6-7} \\
        \bottomrule
    \end{tabular}
    \label{tab:ind_results}
\end{table}
\\
\begin{table}[h]
    \caption{Late fusion ensemble results on NELL-995 (Hits@10)}
    \centering
    \begin{tabular}{@{}llllll@{}}
        \toprule
        & TransE & DistMult & ComplEx & RotatE & GraIL \\
        \midrule
        %  \cmidrule(lr){2-3} \cmidrule(lr){4-5} \cmidrule(lr){6-7} \\
        % \addlinespace[-12pt]
        T & 98.50 &	98.32 &	98.43 &	98.54 &	98.45 \\
        D & & 95.68 &	95.92 &	97.77 &	97.79 \\
        C & & & 95.43 &	97.88 &	97.86 \\
        R & & & & 98.09 &	98.24 \\
        G & & & & & 94.54 \\
        % \addlinespace[10pt]
        % \cmidrule(lr){2-3} \cmidrule(lr){4-5} \cmidrule(lr){6-7} \\
        \bottomrule
    \end{tabular}
    \label{tab:ind_results}
\end{table}

\begin{table}[h]
    \caption{Late fusion ensemble results on FB15k-237 (Hits@10)}
    \centering
    \begin{tabular}{@{}llllll@{}}
        \toprule
        & TransE & DistMult & ComplEx & RotatE & GraIL \\
        \midrule
        %  \cmidrule(lr){2-3} \cmidrule(lr){4-5} \cmidrule(lr){6-7} \\
        % \addlinespace[-12pt]
        T & 98.87 &	98.96 &	99.05 &	98.87 &	98.71 \\
        D & & 98.67 &	98.84 &	98.86 &	98.41 \\
        C & & & 98.88 &	98.94 &	98.64 \\
        R & & & & 98.81 &	98.66 \\
        G & & & & & 75.87 \\
        % \addlinespace[10pt]
        % \cmidrule(lr){2-3} \cmidrule(lr){4-5} \cmidrule(lr){6-7} \\
        \bottomrule
    \end{tabular}
    \label{tab:ind_results}
\end{table}

\newpage
  \begin{table*}[tp]
    \caption{Statistics of inductive benchmark datasets}
    \centering
    \begin{tabular}{@{}lllllllllll@{}}
        \toprule
        & & \multicolumn{3}{c}{WN18RR} & \multicolumn{3}{c}{FB15k-237} & \multicolumn{3}{c}{NELL-995} \\
        & & \#relations & \#nodes & \#links & \#relations & \#nodes & \#links & \#relations & \#nodes & \#links \\
        \midrule
        %  \cmidrule(lr){2-3} \cmidrule(lr){4-5} \cmidrule(lr){6-7} \\
        % \addlinespace[-12pt]
        \multirow{2}{*}{v1} & train & 9 & 2746 & 6678 & 183 & 2000 & 5226 & 14 &10915 & 5540 \\
        & \textit{ind-test} & 9 & 922 & 1991 & 146 & 1500 & 2404 & 14 & 225 & 1034 \\
        \midrule
        \multirow{2}{*}{v2} & train & 10 & 6954 & 18968 & 203 & 3000 & 12085 & 88 & 2564 & 10109 \\
        & \textit{ind-test} & 10 & 2923 & 4863 & 176 & 2000 & 5092 & 79 & 4937 & 5521 \\
        \midrule
        \multirow{2}{*}{v3} & train & 11 & 12078 & 32150 & 218 & 4000 & 22394 & 142 & 4647 & 20117 \\
        & \textit{ind-test} & 11 & 5084 & 7470 & 187 & 3000 & 9137 & 122 & 4921 & 9668 \\
        \midrule
        \multirow{2}{*}{v4} & train & 9 & 3861 & 9842 & 222 & 5000 & 33916 & 77 & 2092 & 9289 \\
        & \textit{ind-test} & 9 & 7208 & 15157 & 204 & 3500 & 14554 & 61 & 3294 & 8520 \\
        % \addlinespace[10pt]
        % \cmidrule(lr){2-3} \cmidrule(lr){4-5} \cmidrule(lr){6-7} \\
        \bottomrule
    \end{tabular}
    \label{table:data-stats}
\end{table*}

\newpage
\section{Inductive Graph Generation}
The inductive train and test graphs examined in this paper do not have overlapping entities. To generate the train graph we sampled several entities uniformly to serve as roots then took the union of the k-hop neighborhoods surrounding the roots. We capped the number of new neighbors at each hop to prevent exponential growth. We remove the samples training graph from the whole graph and sample the test graph using the same procedure. The parameters of the above process are adjusted to obtain a series of graphs of increasing size. The statistics of different datasets collected in this manner are summarized in Table \ref{table:data-stats}. Overall, we generate four versions of inductive datasets from each knowledge graph with increasing sizes.

\section{Proof of Theorem \ref{thm:main}}

We restate the main Theorem for completeness.

\begin{theorem}\label{thm:main_2}
Let $\mathcal{R}$ be any logical rule (i.e., clause) on binary predicates of the form:
\small{\begin{equation}\label{eq:abs_rule}
    r_t(X,Y) \leftarrow r_1(X,Z_1) \land r_2(Z_1, Z_2) \land ... \land r_k(Z_{k-1},Y),
\end{equation}}
where $r_t,r_1,...,r_k$ are (not necessarily unique) relations in the knowledge graph, $X,Z_1,...,Z_k,Y$ are free variables that can be bound by arbitrary unique entities.
% , and where we have that
% \begin{equation*}
%     r^*(X,Y) \leftarrow r_1(X,Z) \land r_2(Z,Y)
% \end{equation*}
% for the special case $k=2$.
For any such $\mathcal{R}$ there exists a parameter setting $\Theta$ for a \name\ model with $k$ GNN layers and where the dimension of all latent embeddings are $d=1$ such that 
$$\textrm{score}(u, r^*, v) \neq 0$$
if and only if $r_t(x,y)$ is the head of rule $\mathcal{R}$ and $\exists Z_1,...,Z_k$ where the body of $\mathcal{R}$ is satisfied with $X=u$ and $Y=v$.
\end{theorem}

We prove this Theorem by first proving the following two lemmas.

\begin{lemma}\label{lemma:1}
Given a logical rule $\mathcal{R}$ as in \ref{eq:abs_rule}, we have $r_t$ in the head and $r_i$ is any relation in the body at a distance $i$ from the head. Then the attention weight between any node nodes, $s$ and $t$, connected via relation $r$, $\alpha_{rr_tst}^l$, at layer $l$ can be learnt such that
$$
\alpha_{rr_tst}^l > 0
$$
if and only if $r = r_l$.
\end{lemma}

\xhdr{Proof} For simplicity, let us assume a simpler version of $\alpha_{rr_tst}^l$ as follows
$$
\alpha_{rr_tst}^l = \text{MLP}(r, r_t).
$$
When $r$ and $r_t$ are 1-dimensional scalars (as we assume in Theorem \ref{thm:main}), to prove the stated lemma we need the MLP to learn a decision boundary between the true pair $\mathcal{S}^l: \{(r_l, r_t)\}$ and the induced set of false pairs $\bar{S}^l:\{(r_i, r_j) \enskip \forall (r_i, r_j) \notin \mathcal{S}^l\}$. We also have the flexibility of learning appropriate embeddings of the relations in 1-dimensional space.

This is possible to an arbitrary degree of precision given that MLP with non-linear activation, as is our case, is a universal function approximator \cite{hornik1991approximation}.

\begin{lemma}\label{lemma:2}
For a given rule $\mathcal{R}$ as in \ref{eq:abs_rule} which holds true for a pair of nodes, $X = u$ and $Y = v$, it is possible to learn a set of parameters for a \name\ model such that
$$
\mathbf{h}_t^l > 0
$$
if and only if node $t$ is connected to node $u$ by a path,
$$
r_1(u,Z_1) \land r_2(Z_1, Z_2) \land ... \land r_k(Z_{l-1},t),
$$
of length $l$.
\end{lemma}

\xhdr{Proof} The overall message passing scheme of best performing \name\ model is given by
\footnotesize{\begin{equation}\label{eq:msg}
    \mathbf{h}_t^{l} = \text{ReLU}\left(\mathbf{W}_{self}^l \mathbf{h}_t^{l-1} + \sum_{r=1}^{R} \sum_{s \in \mathcal{N}_{r}(t)} \alpha_{rr_tst}^l \mathbf{W}_r^l \mathbf{h}_s^{l-1}\right)
\end{equation}}

Without loss of generality, we assume all the nodes are labeled with 0, except the node, $u$, which is labeled $1$. Under this node label assignment, for any node $t$, at a distance $d$ from the node $u$, $\mathbf{h}_t^l = 0 \enskip \forall l < d$.

With no loss of generality, also assume $W_r^k = 1, W_{self}^k = 1 \enskip \forall k, r$. With these assumptions, Equation \eqref{eq:msg} simplifies to 
\begin{equation}\label{eq:sim_agg}
    \mathbf{h}_t^{l} = \text{ReLU}\left(\sum_{r=1}^{R} \sum_{s \in \mathcal{N}_{r}(t)} \alpha_{rr_tst}^l \mathbf{h}_s^{l-1}\right).
\end{equation}

We will now prove our Lemma using induction.

\xhdr{Base case} We will first prove the base case for $l=1$, i.e., $\mathbf{h}_t^1 > 0$ if and only if $t$ is connected to $u$ via path $r_1(u, t)$

From Equation \ref{eq:sim_agg}, we have that
$$
\mathbf{h}_t^{1} = \text{ReLU}\left(\sum_{r=1}^{R} \sum_{s \in \mathcal{N}_{r}(t)} \alpha_{rr_t}^1 \mathbf{h}_s^0\right).
$$
According to our simplified node labeling scheme $\mathbf{h}_s^0 \neq 0$ only if $s = u$. And by Lemma \ref{lemma:1}, $\alpha_{rr_tst}^1 > 0$ only if $r = r_1$. Hence, $t$ must be connected to $u$ via relation $r_1$ for $\mathbf{h}_t^1$ to be non-zero.

\xhdr{Induction step} Assume the induction hypothesis is true for some $\lambda$, i.e., $\mathbf{h}_t^\lambda > 0$ if and only if $t$ is connected to source $u$ by a path $r_1(u,Z_1) \land ... \land r_\lambda(Z_{\lambda-1},t)$.

From Equation \ref{eq:sim_agg} we have that $\mathbf{h}_t^{\lambda+1} > 0$ when the following two conditions are simultaneously satisfied.
\begin{enumerate}
    \item $\mathbf{h}_s^\lambda > 0$ for some $s$
    \item $\alpha_{rr_tst}^{\lambda+1} > 0$ for some $r$
\end{enumerate}
As a consequence of our induction hypothesis, Condition 1 directly implies that node $s$ should be connected to source node $u$ by a path $r_1(u,Z_1) \land ... \land r_\lambda(Z_{\lambda-1},s)$.

By Lemma \ref{lemma:1}, Condition 2 implies that $r = r_{\lambda+1}$. This means that node $t$ is connected to node $s$ via relation $r_{\lambda+1}$.

The above two arguments directly imply that $\mathbf{h}_t^{\lambda+1} > 0$ if and only if node $t$ is connected to source node by a path $r_1(u,Z_1) \land ... \land r_\lambda(Z_{\lambda-1},s)\land r_{\lambda+1}(s,t)$.

Hence, assuming the lemma holds true for $\lambda$, we proved that holds it true for $\lambda + 1$. Thus, Lemma 1 is proved by induction.

\xhdr{Proof of Theorem \ref{thm:main}} This is a direct consequence of Lemma \ref{lemma:2}. In particular, without any loss of generality we simplify the final scoring of \name\ to directly be the embedding of the target node $v$ at the last layer $k$, i.e,
$$
\text{score}(u, r_t, v) = \mb{h}^k_v
$$
According to Lemma \ref{lemma:2}, $\mb{h}_v^k$ is non-zero only when $v$ is connected to $u$ by the body of rule $\mathcal{R}$, hence proving the above stated theorem.

\end{document}